%% file: root.tex
\documentclass[letterpaper, 10 pt, conference]{ieeeconf}  

\IEEEoverridecommandlockouts                              

\usepackage{graphics} 
\usepackage{graphicx}
\usepackage{amsmath} 
\usepackage{amssymb}  
\usepackage{tabularray}
\usepackage{color}
\usepackage{xcolor}
\usepackage{cite}
\usepackage[colorlinks=true, linkcolor=red, citecolor=black, urlcolor=blue]{hyperref}
\usepackage{tabularx}
\usepackage{booktabs}

\title{\LARGE \bf
CoVAR: Co-generation of Video and Action for Robotic Manipulation via Multi-Modal Diffusion
}
\definecolor{darkgreen}{RGB}{0,100,0}

\author{
Liudi Yang$^{1}$,
Yang Bai$^{2,3}$,
George Eskandar$^{5}$,
Fengyi Shen$^{4,5}$,\\
Mohammad Altillawi$^{5}$,
Dong Chen$^{5}$,
Ziyuan Liu$^{5\dagger}$\thanks{$\dagger$ Corresponding author},
Abhinav Valada$^{1}$\\[0.5em]
$^{1}$University of Freiburg,
$^{2}$Ludwig Maximilian University of Munich,\\
$^{3}$Munich Center for Machine Learning (MCML),
$^{4}$Technical University of Munich,\\
$^{5}$Huawei Heisenberg Research Center (Munich)
}

\begin{document}

\maketitle
\thispagestyle{empty}
\pagestyle{empty}

\input{tex/abstract}

\input{tex/introduction}

\input{tex/related_work}
\input{tex/method}

\input{tex/experiment}
\input{tex/conclusion}






{
\bibliographystyle{IEEEtran}
\bibliography{IEEEabrv, ref}
}

\end{document}

%% file: tex/abstract.tex
\begin{abstract}

We present a method to generate video–action pairs that follow text instructions, starting from an initial image observation and the robot’s joint states. Our approach automatically provides action labels for video diffusion models, overcoming the common lack of action annotations and enabling their full use for robotic policy learning. Existing methods either adopt two-stage pipelines, which limit tightly coupled cross-modal information sharing, or rely on adapting a single-modal diffusion model for joint distribution that cannot fully leverage pretrained video knowledge. To overcome these limitations, we (1) extend a pretrained video diffusion model with a parallel, dedicated action diffusion model that preserves pretrained knowledge, (2) introduce a Bridge Attention mechanism to enable effective cross-modal interaction, and (3) design an action refinement module to convert coarse actions into precise controls for low-resolution datasets. Extensive evaluations on multiple public benchmarks and real-world datasets demonstrate that our method generates higher-quality videos, more accurate actions, and significantly outperforms existing baselines, offering a scalable framework for leveraging large-scale video data for robotic learning.
\end{abstract}

%% file: tex/introduction.tex
\section{Introduction}
Video diffusion models have shown great promise toward the goal of developing generalist robotic policies. Their ability to simulate physical interactions~\cite{videosimulator}, propose feasible plans~\cite{vlp}, and augment novel data~\cite{dreamgen} through large-scale pretraining contributes significantly to robotic policy learning. However, the absence of direct paired action labels creates a substantial gap, limiting the effective use of newly generated visual data for robotic imitation learning and visual plan execution.

Previous methods (Fig.~\ref{fig:teaser}) address this problem by attaching a learnable policy model to infer robotic actions from generated videos~\cite{unipi,vpp,dreamgen,genieEnvisioner}. This two-stage design has the advantage of leveraging powerful pretrained video diffusion models, thereby reducing the need for large-scale paired video–action data. However, it also faces challenges for action generation: (1) its performance relies heavily on the quality of video generation, and (2) it struggles when only a small portion of the robotic arm is visible in the generated videos. More recently, joint video–action generation has emerged as an active research direction~\cite{pad,uva,uwm}, motivated by the advantages of learning a shared latent space. These approaches strengthen the policy model by leveraging information from co-modeling video and action space. {Nevertheless, they typically require training a joint diffusion model from scratch or adapting a video diffusion model to a shared latent space using a massive amount of data to capture the joint distribution of the two modalities}. This often leads to suboptimal performance when only limited high-quality data are available.
These limitations call for a framework that not only inherits the general visual knowledge of pretrained video diffusion models, as in two-stage approaches, but also facilitates information sharing between video and action modalities, as in joint models.

\input{figure/teaser}

{In this paper, we propose a novel multi-modal diffusion model capable of simultaneously generating both video and robotic actions for manipulation tasks. Unlike previous joint diffusion models, which either adapt the video Diffusion Transformer (DiT)~\cite{dit} to a shared latent space or train a model from scratch, we incorporate a dedicated action DiT model parallel to the video DiT for generating video-action pairs,} 
allowing the video DiT to fully inherit versatile knowledge from pretrained models. This design reduces the difficulty of learning a joint latent space when data is limited. {To facilitate information sharing across modalities, we introduce a Bridge Attention module. Compared with standard attention mechanisms, which use a single set of queries, keys, and values for all modalities, Bridge Attention employs separate queries, keys, and values for each modality. }
 This mechanism strengthens the alignment of actions with video signals, thereby enhancing action representation learning on top of a large-scale pretrained video diffusion model. 
{A key component of our approach is the action decoder, which we design based on a UNet~\cite{unet} architecture. This choice enables richer cross-modal feature representations and leads to more accurate action generation.} To further improve action accuracy on datasets with low-resolution videos, 
we propose a transformer-based action refinement model that conditions on the initial observation image and textual description, transforming coarse actions generated by our model into precise action sequences, thereby enabling more reliable task execution. 

Extensive evaluations on public benchmarks and our self-collected real-world datasets demonstrate that our proposed \textbf{CoVAR} model achieves substantial improvements in aligning predicted actions with generated videos. The quality of the resulting video–action pairs can be explicitly verified through quantitative metrics including action prediction accuracy and video generation fidelity 
confirming the effectiveness of our approach. Notably, in challenging real-world experiments involving fine-grained manipulation tasks such as picking up tiny nuts, screws, and dowels, our method surpasses baseline approaches by a significant margin.

Our contributions can be summarized as follows:
\begin{itemize}
\item {We extend a video diffusion model with a parallel, dedicated action DiT to form a multi-modal framework that simultaneously generates videos and robotic actions for manipulation tasks.}
\item {We propose Bridge Attention mechanism that assigns separate queries, keys, and values to video and action modalities, enabling more effective cross-modal interaction. }
\item {
We design an action refinement model conditioned on text instruction and initial image observation that improves low-resolution datasets by converting coarse actions into precise actions.
}
\item {We conduct extensive experiments on both simulation benchmarks and real-world robotic datasets, demonstrating that our method consistently generates higher-quality videos and more accurate actions than existing baselines.} 
 \end{itemize}


%% file: figure/teaser.tex
\begin{figure}
    \centering
    \includegraphics[width=0.95\linewidth]{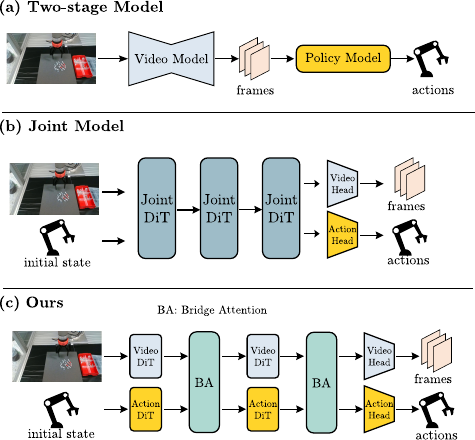}
    \caption{{Architectural comparison with prior methods. }(a) Two-stage Model \cite{vpp,dreamgen,unipi,avdc,genieEnvisioner} (b) Joint Model \cite{uva,uwm,pad} (c) CoVAR. Different from other methods, our framework extends video DiT by attaching a dedicated DiT for action generation, meanwhile allowing the action branch to share information with the pretrained video backbone via our proposed Bridge Attention.    } 

    \label{fig:teaser}
    \vspace{-5pt}
\end{figure}

%% file: tex/related_work.tex
\section{Related Work}

\subsection{Video and Action Generation for Robotics}
Policy learning based on video generation has emerged as a promising direction for building general-purpose agents. Existing approaches can be broadly categorized into two paradigms: \textbf{two-stage} methods and \textbf{joint models} (Fig. \ref{fig:teaser}).

{\parskip=2pt
\noindent\textbf{Two-stage}: 
Two-stage methods first employ a video generation model to predict a robot’s future visual plan, and then attach a policy model conditioned on the generated video or its latent representations. Unipi~\cite{unipi}  was the first to reformulate sequential decision making as a video generation problem, where a lightweight inverse dynamics model is used to infer corresponding actions from videos. Subsequent works have explored different strategies to enhance policy generalization and accuracy. For instance, AVDC~\cite{avdc} integrates depth estimation and optical flow to improve generalization capability. \cite{vpp,genieEnvisioner} leverage intermediate video diffusion latents to learn more accurate implicit inverse dynamics models. 
DreamGen~\cite{dreamgen} pseudo-labels generated actions as supplementary data for more robust policy learning. Vidar~\cite{vidar} extends single-view video diffusion to multi-view generation and introduces a masked policy model to adapt to unseen backgrounds. Enerverse~\cite{enerverse} incorporates sparse context memory for long-term autoregressive action chunk learning. 
\cite{worldvla,gr2} enhance policy learning by equipping VLA models with video generation capabilities. Finally, DiWA~\cite{diwa} and \cite{grounding} utilize generated video states as visual goals and rewards to guide policy exploration. Despite substantial advances, the two-stage action generation model still struggles in complex scenes when the generated video quality is insufficient. {In contrast, our proposed method enables effective information sharing between the two modalities within a multi-modal diffusion model.}}

{\parskip=0pt
\noindent\textbf{Joint Model}: 
Recently, denoising in a joint latent space of video and action has emerged as a powerful paradigm, drawing significant attention for its ability to tightly couple visual prediction and control and thereby enhance visuomotor policy learning. PAD~\cite{pad} integrates prediction and action within the same diffusion transformer (DiT) architecture, merging all modality outputs into a single denoising procedure. UVA~\cite{uva} learns a joint video–action latent representation while employing separate diffusion heads for video and action decoding to improve inference efficiency, with a masked autoencoder further enhancing performance in complex scenarios. 
UWM~\cite{uwm} incorporates both an action diffusion process and a video diffusion process within a unified transformer architecture, enabling flexible representation of policy, forward dynamics, inverse dynamics, and video generation. 
These aforementioned models train a diffusion model from scratch or adapt a pretrained video diffusion model to learn the joint distribution, often leading to suboptimal performance with limited data. 
{Concurrent work~\cite{videogenerator} leverages video generation as a proxy for policy learning by employing two separate denoising UNets, which improves generalization and enables the use of action-free data. In contrast to its one-way information flow from video to action, our Bridge Attention enables bidirectional information exchange between modalities, leading to more effective training.}} 
\input{figure/method_figure}
\subsection{Multi-modal Generation Model}
Multi-modality generation models serve as fundamental building blocks toward general-purpose, multi-task generative systems. Recently, OmniFlow~\cite{omniflow} introduced a multimodal rectified flow formulation to enable any-to-any generation. \cite{unidiffusion,multidiffusion} explicitly fit all relevant distributions within a single model, achieving this with minimal modifications to the original diffusion framework and without additional training or inference overhead.  
\cite{mmdiffusion} was the first to present two coupled denoising autoencoders for joint audio–video generation. 
GEM~\cite{gem} generates paired RGB and depth outputs for enhanced spatial understanding. Finally, \cite{boost} introduces a generative image modeling framework that leverages diffusion models to jointly capture low-level image latents and high-level semantic features from a pretrained self-supervised encoder. These works demonstrate the potential of diffusion models to generate aligned multi-modal content including depth, PCA features, audio, and so on, motivating our investigation into video–action co-generation for more accurate and scalable policy learning.
{However, extending this paradigm to embodied AI requires more careful architectural design, since video and actions are fundamentally different in nature. Unlike modalities such as RGB and depth, they are not spatially aligned and therefore demand tailored mechanisms for effective joint modeling.}



%% file: figure/method_figure.tex
\begin{figure*}[htp]
    \vspace{5pt}
    \centering
    \includegraphics[width=0.9\linewidth]{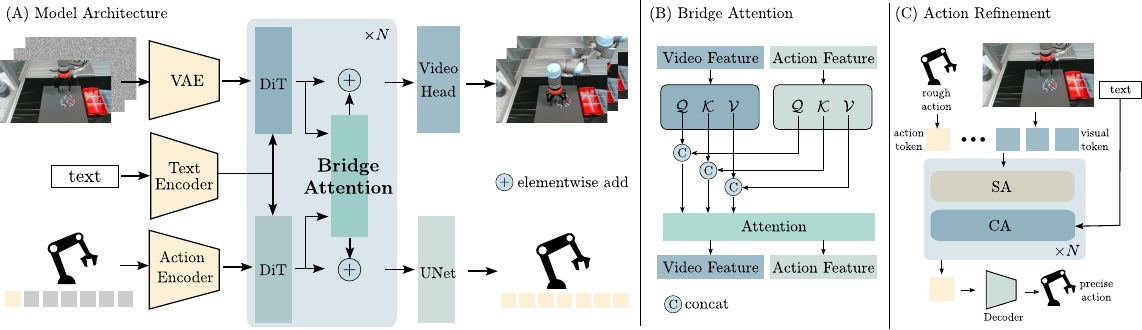}
    \caption{Overview of CoVAR. (A) It is built on a video diffusion backbone with a parallel Action DiT to generate actions. (B) The two modalities interact through Bridge Attention. (C) For low-resolution datasets, an Action Refinement Module is introduced. }
    \label{fig:method}
\end{figure*}

%% file: tex/method.tex
\section{Method}

Given an initial observation image $v_0 \in \mathbb{R}^{3 \times H \times W}$, the robot’s initial joint states $a_0 \in \mathbb{R}^{L}$, and a natural language instruction $c$, we train a multi-modal  diffusion model to generate video frames $v \in \mathbb{R}^{T \times 3 \times H \times W}$ and paired actions $a \in \mathbb{R}^{T \times L}$ that are aligned with the instruction. We first introduce multi-modal rectified flow as a preliminary in Sec.~\ref{sec:rf}. Sec.~\ref{sec:ModelArchitecture} introduces the architecture of our multi-modal diffusion model. In Sec.~\ref{sec:bridge_attention}, we describe the proposed Bridge Attention mechanism. Finally, Sec.~\ref{sec:action_refine} presents the action refinement model, which is designed to improve performance on datasets with lower spatial resolution. The overview is shown in Fig.~\ref{fig:method}.

\subsection{Multi-modal Rectified Flow}
\label{sec:rf}
To generate higher-quality video–action  pairs and accelerate convergence during training, we adopt a multi-modal rectified flow model \cite{rflow,flowmatching}, which simplifies the diffusion process by constructing a straight-line path from noise to data. We consider the joint data distribution $X_0 = \big(x_0^{1}, x_0^{2}\big) \sim \Pi_{data}$, where $x_0^{1}$ denotes the video latent and $x_0^{2}$ denotes the robotic action. Rectified flow learns a flow field $v_t(X_t)$ that maps $X_0$ to
$
X_1 = \big(x_1^{1}, x_1^{2}\big) \sim 
\big(\mathcal{N}(0, \mathbf{I}_{d_1}), \; \mathcal{N}(0, \mathbf{I}_{d_2})\big),
$
where $d_1$ and $d_2$ denote the dimensionalities of the first and second modalities, respectively. 
This mapping is defined through an ordinary differential equation (ODE), which is approximated by linear interpolation:
\begin{equation}
\label{eq:diff_equation}
\frac{dX_t}{dt} = X_1 - X_0, \quad t \in [0,1].
\end{equation}



A neural network $v_{\theta}=(v_{\theta}^{1},v_{\theta}^{2})$ models this mutli-modal vector field, 
and it is trained by minimizing the objective:
\begin{equation}
\label{eq:modality_loss}
    L = \| x_1^{1}-x_0^{1}-v_{\theta}^{1} \|_2 +  \|  x_1^{2}-x_0^{2}-v_{\theta}^{2} \|_2.
\end{equation}

At inference time, samples are generated by integrating the learned flow in Eqn. (\ref{eq:diff_equation}).

\subsection{Model Architecture}
\label{sec:ModelArchitecture}
Building on the pretrained video diffusion model Open-Sora~\cite{opensora}, we extend it to jointly generate video–action pairs, which is beneficial for both robotic manipulation data augmentation and standalone policy learning. To preserve the general knowledge acquired from large-scale datasets of video diffusion models, we attach a parallel action DiT module alongside the video DiT backbone, in contrast to prior approaches that employ a single joint DiT to model both modalities. The use of two dedicated DiT modules for video and action, respectively, enables improved performance under limited data conditions. Given the relatively low dimensionality of action data, we avoid the costly training of a VAE (commonly used in latent diffusion models to reduce the denoising process into a lower-dimensional latent space) and instead use a lightweight MLP encoder to obtain action embeddings. 

The action DiT is conditioned on text instruction $c$ through cross-attention, forming a symmetric counterpart to the video DiT. Information exchange between action feature and video feature is enabled by the proposed Bridge Attention mechanism (Sec.~\ref{sec:bridge_attention}).  The Bridge Attention mechanism mitigates interference between the two modalities, enabling them to reconcile and integrate more effectively. 
Empirically, we find that the choice of action decoder plays a critical role in both accuracy and convergence speed during training. In contrast to prior work that relies on MLPs or ResNets, we employ a UNet \cite{unet}, which yields more accurate and better-aligned action generation. The overall model is optimized by minimizing the loss $L$ defined in Eqn.(\ref{eq:modality_loss}). 

\subsection{Bridge Attention}
\label{sec:bridge_attention}
{Prior works employ a joint DiT for the two modalities, performing early fusion by concatenating video and action tokens and feeding them into the same DiT, which enables intrinsic information sharing between them. However, it also requires the joint DiT to adapt pretrained video diffusion knowledge into a unified latent space, resulting in suboptimal performance with limited expert demonstrations.} 

{In our pipeline, we build on the design concept of \cite{omniflow}, which facilitates cross-modal interactions, and extend it as Bridge Attention mechanism. This extension adapts the framework to the action modality within a DiT architecture, enabling more effective information exchange between the video and action streams.}
Bridge Attention allows each modality to maintain its own dedicated representation space, while still enabling cross-modal communication through attention. Specifically, we parameterize queries ($q_1,q_2$), keys($k_1,k_2$), and values ($v_1,v_2$) separately for video features $f_v \in \mathbb{R}^{B \times N_v \times C} $ and action features $f_a \in \mathbb{R}^{B \times N_a \times C} $, where $B$ is the batch size, $N_v$ and $N_a$ are the token number of video and action, $C$ is the feature dimension. We then concatenate them  to compute attention and split them back into video features and action features:
\begin{equation}
    \begin{bmatrix} f_v \\ f_a \end{bmatrix} = \mathrm{Attention}\Big(
        \begin{bmatrix} q_1 \cdot f_v \\ q_2 \cdot f_a \end{bmatrix},
        \begin{bmatrix} k_1 \cdot f_v \\ k_2 \cdot f_a \end{bmatrix},
        \begin{bmatrix} v_1 \cdot f_v \\ v_2 \cdot f_a\end{bmatrix}
    \Big).
    \label{eq:bridge_attention}
\end{equation}

{This design offers a key advantage over traditional attention mechanisms: it mitigates cross-modal interference, where one modality can overwhelm or distort the representation of the other.  }
In standard self-attention, video and action tokens are concatenated and processed together using the same queries, keys, and values, 
which can cause one modality to dominate the other and lead to misalignment. 
Bidirectional cross-attention allows information exchange between modalities but does not preserve each modality’s internal structure, which can lead to interference and degrade performance when high-quality data are limited.
Bridge Attention maintains separate queries, keys, and values for each modality while enabling joint attention over concatenated tokens, striking a balance between intra-modal consistency and inter-modal interaction.  As shown in our ablation studies (Tab.~\ref{tab:ablation}), this mechanism consistently improves both video generation quality and action accuracy, demonstrating its effectiveness in multi-modal modeling. 

\subsection{Action Refinement Model}
\label{sec:action_refine}
To generate high-resolution content, video diffusion models are typically fine-tuned on high-resolution videos (e.g., 480p or 720p) during the final stage of general pretraining. However, many robotic datasets provide only low-resolution data (e.g., 128p). This resolution mismatch makes it challenging to generate accurate trajectories and often reduces task success. In such low-resolution scenarios, our model is capable of generating video along with rough actions, but these actions may lack the precision required for task execution.

To address this limitation, we introduce an action refinement model that converts rough actions into high-precision actions, conditioned on the initial image and text instruction, which is shown in Fig. \ref{fig:method}(C). Specifically, the refinement model first obtains action tokens and image tokens by passing the input coarse actions and initial image observation through their respective embedding networks. It then concatenates the action tokens with the corresponding image tokens and processes them through a self-attention module, allowing the model to aggregate visual information into the action tokens. This is followed by a cross-attention module conditioned on the text description $c$ to incorporate instruction guidance. {Finally, the decoder maps the intermediate features to the appropriate output dimensions, yielding the refined robotic actions.}


%% file: tex/experiment.tex
\input{figure/video_trajectory}

\section{Experiments}

In this section, we present experiments to evaluate the efficacy of our approach in the following aspects:  
\begin{itemize}
    \item Improved video generation quality (Sec.~\ref{sec:video_quality}).  
    \item Higher success rate of action generation (Sec.~\ref{sec:action_quality}).  
\end{itemize}

\subsection{Setup and Baselines}
{\parskip=0pt
\noindent\textbf{Datasets}: We evaluate our method on two public datasets: Calvin~\cite{calvin} and Libero90~\cite{libero} as well as on our self-collected real-world dataset. The Calvin dataset contains approximately 20k teleoperated demonstrations paired with text instructions, with videos recorded at a resolution of $200 \times 200$. The Libero90 dataset consists of 90 tasks spanning diverse scenarios, with 50 expert demonstrations provided for each task. Since the video resolution is only $128 \times 128$, we apply our proposed action refinement model to this dataset. This module does not need to be applied to other higher-resolution datasets. For real-world evaluation, we collected a dataset of 1K demonstrations covering tasks such as bowl stacking, and the picking and placing of screws, nuts, and dowels.}

{\parskip=2pt
\noindent\textbf{Training Details}: We use OpenSora-1.2 as our codebase. The model parameter is 1.4B (video diffusion 1.1B + our new module 0.3B). We sample 35 frames from the aforementioned dataset for training.  {We co-train the two modalites on the foundation of pretrained video diffusion model. } 
For the real dataset, the video resolution is $180 \times 320$ for fast convergence and inference speed. The training takes about 1 day on 4 GPUs. For the action refinement model on the Libero90 dataset, we finetune our action refinement  model with 450 video-action pairs.}

{\parskip=1pt
\noindent\textbf{Real Rollout Details}: We infer the video–action pair of  35 frames from the current observation and the robot’s current joint state via ROS communication. The sampling step of rectified flow is set to 30, resulting in a generation time of approximately 4 seconds for one sequence. Since the robot operates at a control frequency of 100 Hz, we interpolate the generated open-loop action sequence to match this control frequency.}

{\parskip=2pt
\noindent\textbf{Baseline}: We compare our method against joint models and two-stage models in terms of video quality and action accuracy.
\begin{itemize}
    \item UVA\cite{uva}: A\textbf{ joint DiT-based} model that learns joint video-action representations with masked training strategy.
    \item UWM\cite{uwm}: A \textbf{joint DiT-based} model that enables scalable robot learning from both annotated actions and large-scale action-free video data through policy learning, dynamics modeling, and video generation.
    \item PAD \cite{pad}: A \textbf{joint DiT-based} framework that unifies video prediction and robotic action generation within a joint denoising process.
    \item Unipi \cite{unipi}: \textbf{A two-stage} model that first generates future video frames to plan actions, and then uses a learnable inverse dynamics model to convert video plans into robot actions.
    \item RoboEnvision \cite{roboenvision}:  A \textbf{two-stage} pipeline that first generates robotic manipulation videos conditioned on high-level goals. Second, a lightweight policy model regresses robot joint states as control commands.
\end{itemize}}

\input{figure/video_comparison}
\subsection{Evaluation of Video Quality}
\label{sec:video_quality}

{\parskip=0pt
\noindent\textbf{Metrics}: We report PSNR, SSIM, LPIPS, and FVD as metrics for evaluating video quality. During training, the ground-truth data are partitioned into a validation set, which is used to compute these metrics.}

{\parskip=2pt
\noindent\textbf{Quantitative Results}: We evaluate 100 generated videos on the Calvin dataset and 180 on the Libero90 dataset, calculating the metrics reported in Tab. \ref{tab:video_comparision}. On both datasets, our method outperforms three joint-model baselines across all four metrics. Compared to the pure video diffusion model used in the two-stage approach, OpenSora‑1.2, our metrics are comparable, indicating that the inclusion of the action modality does not degrade the quality of generated videos.}

\input{table/video_comparison_table}

{\parskip=2pt
\noindent\textbf{Qualitative Results}: 
As shown in the qualitative results of Fig. \ref{fig:video_trajectory}, our co-generation model is capable of producing high-fidelity videos while simultaneously generating accurate trajectories across diverse datasets, including both simulated and real-world demonstrations on different robotic platforms.
We also compare our video generation results with those of the joint-model baselines in Fig. \ref{fig:video_comparison}. We observe that our approach generates videos with fewer artifacts, greater visual clarity, and more consistent robotic arm and objects.}

\subsection{Evaluation of Action Success Rate}
\label{sec:action_quality}

{\parskip=0pt
\noindent\textbf{Calvin Dataset}: 
We categorize the CALVIN tasks into five groups: opening/closing drawers, opening/closing cabinets, turning lights on/off via handles and buttons, picking objects, and pushing objects. We train with 20k videos from setting ABCD, and we randomly generate 200 novel test scenes as rollouts. The quantitative results are presented in Tab.~\ref{tab:calvin}. Compared with the baselines, our method achieves a substantially higher success rate, demonstrating its effectiveness for accurate action generation. In addition, we provide a sample rollout in the second row of Figure~\ref{fig:video_comparison}.} 

\input{table/calvin_table}

{\parskip=2pt
\noindent\textbf{Libero90 Dataset}: 
We categorize the 90 tasks into three groups: pick-and-place, opening/closing drawers, microwaves, and ovens, and compositional tasks. The total number of rollouts is 450. The quantitative results are reported in Tab.~\ref{tab:libero}. Compared with other baselines, our model achieves a significant improvement, aided by the action refinement module. As shown in Fig.~\ref{fig:action_refinement}, the raw actions generated by our model provide a coarse motion plan for the robotic arms. However, for operations requiring fine precision, such as grasping small objects, these raw actions alone are insufficient. With the addition of the action refinement module, the generated trajectories become precise enough to successfully complete such tasks.}
\input{table/libero_table}
\input{figure/action_refinement}

{\parskip=0pt
\noindent\textbf{Real Dataset}:  We evaluate the success rates of three manipulation tasks using the UR5 platform. Notably, the objects involved are very small, requiring highly accurate action generation. As shown in Tab.~\ref{tab:real}, our model achieves the highest success rate compared with the baselines, demonstrating the effectiveness of our approach. Additionally, we present consecutive pick-and-place rollouts in Fig.~\ref{fig:video_rollout}. The real-world rollouts closely resemble our generated videos, highlighting the strong alignment between video and action generation. These results indicate that our model can cogenerate video and action that are both visually coherent and physically executable, even under challenging conditions requiring fine-grained control.}

\input{table/real_table}
\input{figure/video_rollout}

\subsection{Ablation Study}
For a comprehensive understanding of each component in our pipeline, we design experiments on our self-collected dataset, evaluating the effectiveness of the individual submodules.
\input{figure/ablation}
\input{table/ablation_study}

{\parskip=2pt
\noindent\textbf{Bridge Attention}: We first remove our proposed Bridge Attention and replace it with self-attention and cross-attention, respectively, for sharing information between the two modalities. As shown in Tab.~\ref{tab:ablation}, removing the Bridge Attention results in a decline in both video quality and action accuracy, demonstrating that the Bridge Attention effectively mitigates interference between the two modalities. This suggests that standard self-(SA) or cross-attention (CA) mechanisms alone are insufficient for modeling the complex interactions between video and action streams with limited data. The Bridge Attention appears to provide a structured pathway for information exchange, allowing the model to retain modality-specific features while still integrating relevant cross-modal cues, which is critical for generating both high-fidelity videos and accurate actions simultaneously.}

{\parskip=2pt
\noindent\textbf{UNet}: 
We remove the UNet-based action head and replace it with the ResNet architecture used in \cite{roboenvision}. Experimental results indicate that the UNet contributes not only to more accurate action generation but also to improved video quality. This improvement can be attributed to the strong coupling between the action and video streams: more precise action predictions guide the video decoder to produce motion that better aligns with the intended behavior. The UNet’s multi-scale feature processing enables hierarchical representation of temporal action sequences, allowing finer-grained motion dynamics to be captured in both modalities. As a result, the model produces smoother, more realistic video sequences and more reliable robotic actions.}

{\parskip=2pt
\noindent\textbf{Only Action DiT}: We remove the video DiT from our model and train a single-modal diffusion model to generate actions conditioned solely on the text instruction and initial image. The observed decline in success rate demonstrates that the video modality provides crucial information and pretrained knowledge to the action DiT module. This underscores the benefit of multi-modal co-generation, as jointly leveraging video and action streams allows the model to capture complementary cues, resulting in more accurate and robust robotic manipulation.}

%% file: figure/video_trajectory.tex
\begin{figure*}[htbp]
    \centering
    \includegraphics[width=0.9\linewidth]{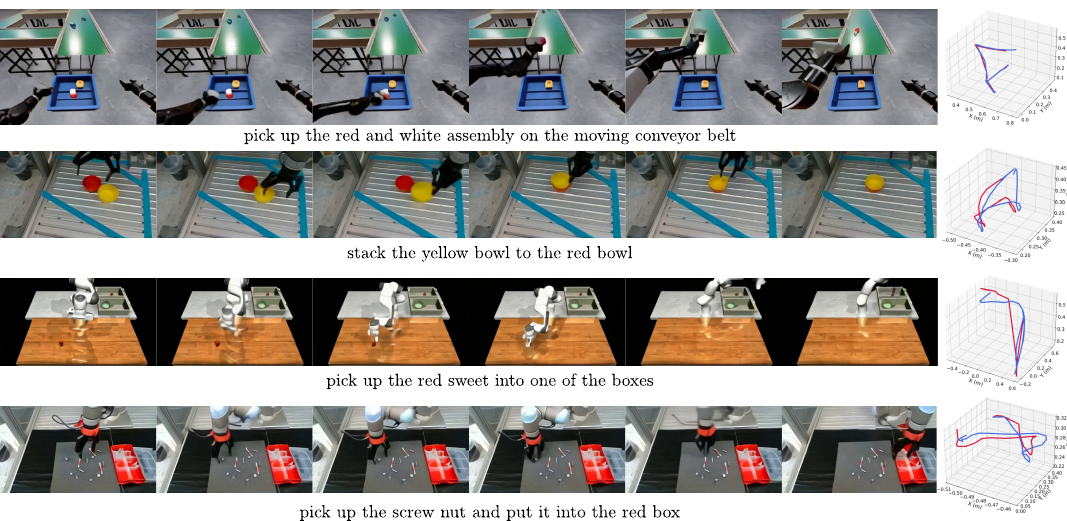}
    \caption{Visualization of generated video-action pair. \textcolor{red}{Red} lines denote groundtruth actions as reference. \textcolor{blue}{Blue} lines denote our generated actions. The generated videos align well with the text instructions, and the paired actions closely match the reference ground truth to achieve the tasks.}
    \label{fig:video_trajectory}
    \vspace{-5pt}
\end{figure*}

%% file: figure/video_comparison.tex
\begin{figure*}
   \vspace{5pt}
    \centering
    \includegraphics[width=\linewidth]{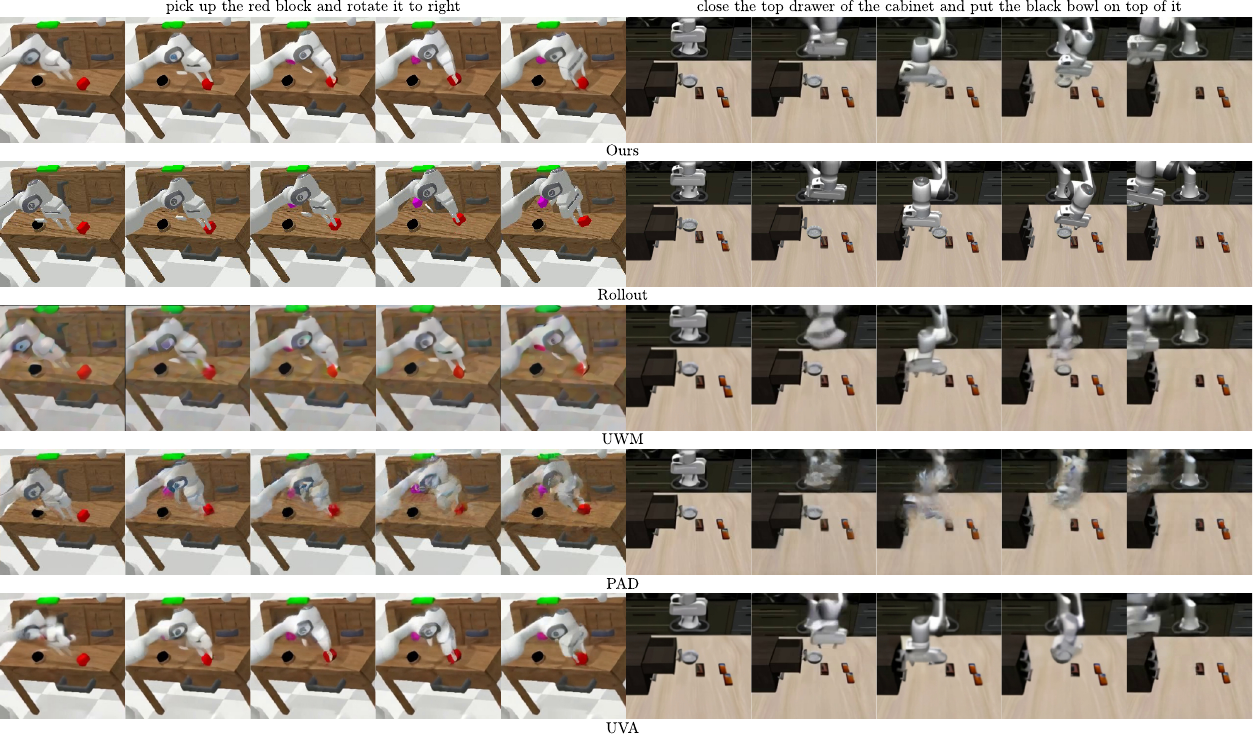}
    \caption{Comparison of generated videos with baselines. In comparison to other baselines, our model generates video content of objects and robotic arms with reduced artifacts, yielding clearer and more realistic results. The rollout shows strong alignment between the generated video and the corresponding action. }
    \label{fig:video_comparison}
\end{figure*}

%% file: table/video_comparison_table.tex
\begin{table}[t]
\centering
\footnotesize
\caption{Quantitative result of video generation quality}
\label{tab:video_comparision}
\setlength{\tabcolsep}{2pt}
\begin{tabular}{lcccccccc}
\toprule
        & \multicolumn{4}{c}{Calvin} & \multicolumn{4}{c}{Libero90} \\
\cmidrule(lr){2-5} \cmidrule(lr){6-9}
        & PSNR & SSIM & LPIPS & FVD & PSNR & SSIM & LPIPS & FVD \\
\midrule
UVA \cite{uva}        & 19.01 & 0.758 & 0.180 & 97.90  & 19.57 & 0.716 & 0.154 & 86.21 \\
PAD \cite{pad}        & 18.72 & 0.734 & 0.174 & 83.40  & 19.65 & 0.781 & 0.218 & 98.39 \\
UWM \cite{uwm}        & 18.04 & 0.730 & 0.181 & 85.85  & 19.87 & 0.735 & 0.212 & 87.83 \\
OpenSora \cite{opensora} 
                      & 19.60 & \textbf{0.768} & 0.171 & \textbf{61.00} 
                      & \textbf{20.18} & 0.817 & 0.156 & \textbf{63.33} \\
Ours                  & \textbf{19.95} & 0.766 & \textbf{0.156} & 72.42 
                      & 20.09 & \textbf{0.826} & \textbf{0.143} & 70.64 \\
\bottomrule
\end{tabular}%
\end{table}

%% file: table/calvin_table.tex


\begin{table}
\centering
\caption{Success Rate on Calvin Dataset}
\label{tab:calvin}
\begin{tabular}{lccccc}
\toprule
Calvin & Drawer & Cabinet & Light & Pick & Push \\
\midrule
UVA \cite{uva}   & 0.875 & 0.667 & 0.711 & 0.758 & 0.785 \\
UWM \cite{uwm}   & 0.813 & 0.733 & 0.644 & 0.576 & 0.714 \\
PAD \cite{pad}   & 0.781 & 0.467 & 0.489 & 0.485 & 0.642 \\
Unipi \cite{unipi} & 0.469 & 0.267 & 0.289 & 0.182 & 0.452 \\
Ours             & \textbf{1.000} & \textbf{0.800} & \textbf{0.867} & \textbf{0.909} & \textbf{0.929} \\
\bottomrule
\end{tabular}%
\end{table}

%% file: table/libero_table.tex



\begin{table}
\centering
\caption{Success rate on Libero90}
\label{tab:libero}
\begin{tabularx}{0.95\linewidth}{l *{3}{>{\centering\arraybackslash}X}}
\toprule
Method & Pick & Open \& Close & Combination \\
\midrule
UVA \cite{uva}            & 0.676 & 0.640 & 0.489 \\
UWM \cite{uwm}            & 0.606 & 0.600 & 0.400 \\
PAD \cite{pad}            & 0.625 & 0.480 & 0.355 \\
Ours (w/o refinement)     & 0.592 & 0.520 & 0.422 \\
Ours                      & \textbf{0.873} & \textbf{0.860} & \textbf{0.711} \\
\bottomrule
\end{tabularx}
\end{table}

%% file: figure/action_refinement.tex
\begin{figure}
    \centering
    \includegraphics[width=\linewidth]{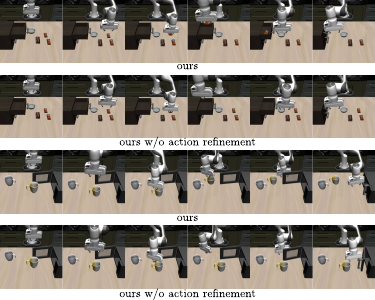}
    \caption{Rollout comparison between our model and the variant without action refinement. The model without action refinement produces coarse actions that merely reflect the general trend of the task but remain imprecise; action refinement enhances precision and enables successful completion.}
    \label{fig:action_refinement}
    \vspace{-5 pt}
\end{figure}

%% file: table/real_table.tex


\begin{table}
\vspace{5pt}
\centering
\caption{Success Rate on Real Dataset}
\label{tab:real}
\begin{tabular}{lccc}
\toprule
            & Nut & Screw & Dowel \\
\midrule
Unipi \cite{unipi}            & 0.00 & 0.06 & 0.02 \\
RoboEnvision \cite{roboenvision} & 0.04 & 0.10 & 0.12 \\
Ours                          & \textbf{0.64} & \textbf{0.74} & \textbf{0.70} \\
\bottomrule
\end{tabular}%
\end{table}

%% file: figure/video_rollout.tex
\begin{figure}[htbp]
    \centering
    \includegraphics[width=\linewidth]{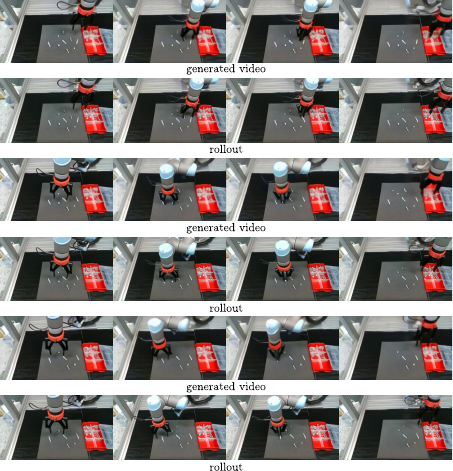}
    \caption{Generated videos and rollouts from real-world experiments. The consecutive success of picking and placing demonstrates high-precision action generation and strong video–action alignment.}
    \label{fig:video_rollout}
    \vspace{-5pt}
\end{figure}

%% file: figure/ablation.tex
\begin{figure}[htbp]
    \vspace{5pt}
    \centering
    \includegraphics[width=\linewidth]{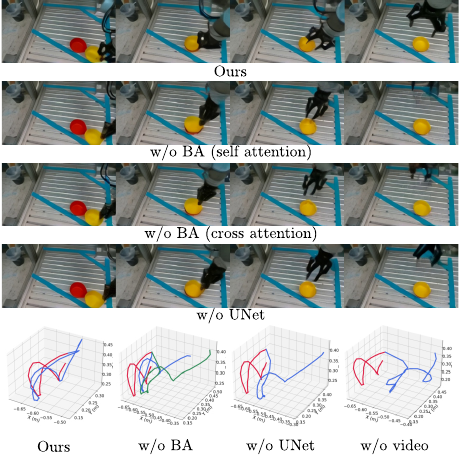}
    \caption{Qualitative results of ablation study. \textcolor{red}{Red} lines denotes groundtruth action. \textcolor{blue}{Blue} lines denotes action generation. In the plotted trajectory of w/o (BA), the \textcolor{blue}{blue} line denotes self-attention and the \textcolor{darkgreen}{green} line denotes cross attention.}
    \label{fig:ablation}
    \vspace{-8pt}
\end{figure}

%% file: table/ablation_study.tex

\begin{table}[t]
\centering
\caption{Ablation Study for Video Generation Quality and Action success rate}
\label{tab:ablation}
\begin{tabular}{lccccc}
\toprule
                  & PSNR   & SSIM   & LPIPS  & FVD    & Rate \\
\midrule
w/o BA (SA) & 16.83 & 0.693 & 0.255 & 137.66 & 0.32 \\
w/o BA (CA) & 16.56 & 0.645 & 0.263 & 145.26 & 0.20 \\
w/o UNet    & 16.85 & 0.690 & 0.255 & 141.62 & 0.24 \\
w/o video   & -     & -     & -     & -      & 0.08 \\
Ours        & \textbf{17.67} & \textbf{0.736} & \textbf{0.238} & \textbf{133.89} & \textbf{0.68} \\
\bottomrule
\end{tabular}%
\end{table}

%% file: tex/conclusion.tex
\section{Conclusion}
In this paper, we propose a multi-modal diffusion model for co-generating video and action in robotic manipulation tasks. Our method preserves video knowledge through a dedicated action DiT for modeling action distributions. To enhance cross-modal interaction, we introduce Bridge Attention, which facilitates more efficient information exchange between the two modalities. The action refinement module further improves performance on datasets with lower resolution. Extensive experimental results show that our approach achieves superior video quality compared with baselines on two public benchmarks. Moreover, it consistently improves the success rate of action generation across both benchmark datasets and real-world experiments. 

{
While our framework demonstrates strong performance, a key limitation is that it operates only on monocular videos,  which restricts its ability to capture full 3D scene geometry. Incorporating 3D perception from foundation models could provide more accurate spatial understanding for both video and action learning.
}